# INCORPORATING LUMINANCE, DEPTH AND COLOR INFORMATION BY A FUSION-BASED NETWORK FOR SEMANTIC SEGMENTATION

*Shang-Wei Hung[1,2]*   *Shao-Yuan Lo[1]*   *Hsueh-Ming Hang[1]*

[1] National Chiao Tung University   [2] UC San Diego

{shangwei.eecs02, sylo2.eecs02}@g2.nctu.edu.tw, hmhang@nctu.edu.tw

**ABSTRACT**

Semantic segmentation has made encouraging progress due to the success of deep convolutional networks in recent years. Meanwhile, depth sensors become prevalent nowadays; thus, depth maps can be acquired more easily. However, there are few studies that focus on the RGB-D semantic segmentation task. Exploiting the depth information effectiveness to improve performance is a challenge. In this paper, we propose a novel solution named LDFNet, which incorporates Luminance, Depth and Color information by a fusion-based network. It includes a sub-network to process depth maps and employs luminance images to assist the depth information in processes. LDFNet outperforms the other state-of-art systems on the Cityscapes dataset, and its inference speed is faster than most of the existing networks. The experimental results show the effectiveness of the proposed multi-modal fusion network and its potential for practical applications.

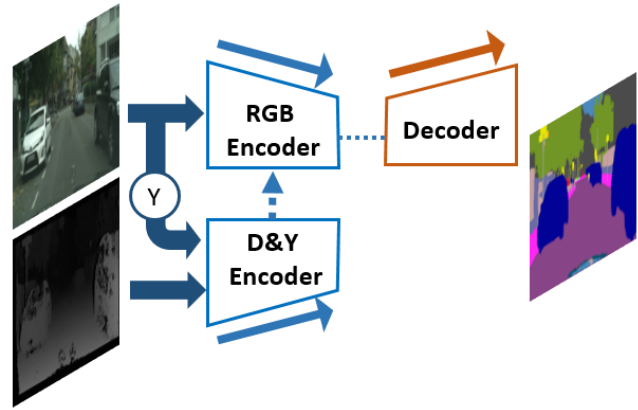

Figure 1: Flowchart of the proposed semantic segmentation system. **Y**: luminance information.

## 1. INTRODUCTION

Because of the success of deep convolutional neural networks (CNNs) in recent years, researchers have made a breakthrough in semantic segmentation. FCN [12] is a pioneer, then SegNet [1], DeepLab [2] and PSPNet [20] are proposed successively. Although these networks show outstanding performance, their computational cost is generally considered too high to be widely deployed. On the other hand, ENet [13] is proposed for low complexity, but its accuracy is much sacrificed. Afterward, ERFNet [14] combines the efficiency of the factorized convolution and the capability of the Non-Bottleneck [7] for better trade-off between accuracy and computational efficiency, but there is still room for further improvement.

More recently, DenseNet [4] introduces a dense connection concept that connects each layer to all the other layers in a feed-forward manner. This strategy reinforces the information propagation and decreases the model complexity. This design is also applicable to the segmentation systems.

Nowadays depth sensors such as Kinect are quite affordable, so RGB-D semantic segmentation is an emerging topic. Typically, because the depth map edges are aligned with RGB image contours, the depth values of objects tend to be uniform or varying gradually along a spatial axis. Therefore, the depth information can be used as a good indicator of objects [17]. The depth maps can thus be treated as complementary data to RGB images, but it is a challenge to extract the complementary information from the depth maps effectively. One simple way is stacking a depth map with a RGB image to form 4 input channels to a CNN, but the attempts so far are not yet successful to exploit the desirable information from depth data complementary to that of RGB data. Gupta et al. [5] introduce the HHA encoding to represent the depth information, yet this transformation does not provide extra useful information than the original depth data itself. FuseNet [6] processes the depth maps by a fusion-based network that feeds the RGB images and the depth maps into two separate sub-networks respectively, then fuses their features together. Even though making some improvements, it increases considerably the number of parameters and the amount of computation.

In this paper, we propose a new solution for RGB-D semantic segmentation, which incorporates both the *Luminance* and *Depth* information by a *Fusion*-based

---

network, named LDFNet. It exploits the information embedded in the depth map by a two-branch architecture similar to that of FuseNet, but we adopt the ERFNet structure as a backbone for the RGB branch, so-called *RGB Encoder* and *Decoder* due to its high efficiency, then we design a new structure for the depth branch (see Figure 1). Our depth branch accepts the notion of the dense connectivity to process the depth maps more efficiently, so that the entire network complexity would not increase too high with the extra depth inputs. Furthermore, we add a dense block at the early stage of the depth branch to purposely extract the boundary and contour features from the depth map.

Because capturing the depth information accurately is a difficult task, the current popular depth sensors cannot provide high quality and high definition depth maps. The captured depth maps are typically at low resolution and have defects such as strong noises and wide occlusion regions. These defects may lead to the poor performance of the depth branch if it works alone by using the captured depth maps only. As a result, inspired by [11,18] that uses the luminance information (Y) for depth map enhancement, we include the luminance images as an input to the depth branch. That is, the luminance images derived from the RGB inputs are stacked with the depth channel in the depth branch to enhance its capability, and thus our depth branch is called *D&Y Encoder*. To the best of our knowledge, we are the first that proposes the D&Y method for RGB-D semantic segmentation. The proposed LDFNet achieves very competitive results in terms of both accuracy and complexity efficiency compared to the other state-of-the-art methods on the Cityscapes dataset [3].

## 2. METHOD

The entire architecture of the proposed LDFNet is shown in Figure 2. It consists of RGB Encoder, Decoder, and D&Y Encoder. In the following paragraphs, we will discuss the details and the reasons behind our network design choices.

### 2.1. RGB Encoder and Decoder

We adopt the network architecture proposed in ERFNet [14] as our network backbone in RGB Encoder and Decoder because of its good performance in considering both reliability (accuracy) and efficiency (complexity). ERFNet is composed of Non-bottleneck-1D by the Non-bottleneck suggested in ResNet [7]. The difference between these two is that each convolutional kernel of the Non-bottleneck is factorized into two one-dimensional convolutional kernels. To be more specific, each $3\times3$ kernel is replaced by a $3\times1$ and a $1\times3$ kernels, and thus the number of parameters can be decreased.

Feature map downsampling makes the receptive fields wider and thus can extract a larger size of contextual representations, but it may also lose detailed spatial information that is especially crucial for semantic segmentation. Rather than overly downsampling the feature maps, compared to SegNet [1] (five downsampling operations in total), ERFNet achieves a better balance by using three Downsampler Blocks. In order to enlarge the receptive fields without additional parameters and computation, the dilated convolutions [2,19] with different rates are interweaved in certain layers.

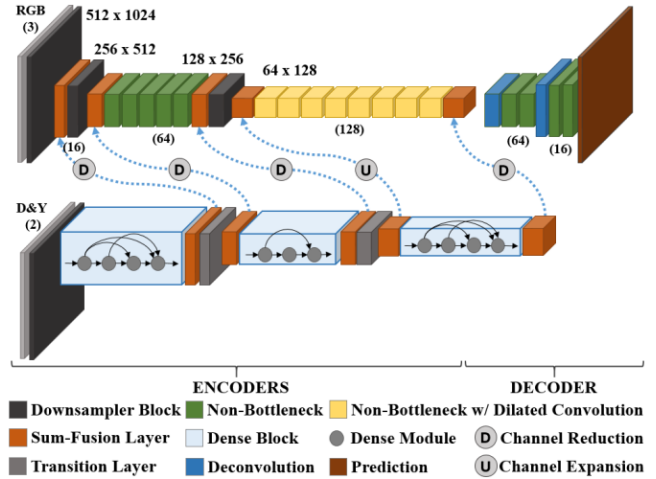

Figure 2: The proposed LDFNet architecture. The numbers in parentheses represent the number of channels.

For the decoder, instead of using the max-unpooling layers introduced in SegNet [1], ERFNet chooses the deconvolution filter for restoring the feature maps to the original resolution.

### 2.2. D&Y Encoder

FuseNet [6] uses two identical architectures for its two encoders. By contrast, our second branch, D&Y Encoder, has a different structure. Because DenseNet [4] is believed to have a much higher efficiency without sacrificing the accuracy, our D&Y Encoder adopts the notion of dense connectivity to enhance the information flow from the earlier layers to the latter layers.

Compared to RGB Encoder, each Non-Bottleneck in the D&Y Encoder is replaced by a dense module. The dense module begins with a $1\times1$ convolution layer for channel reduction to improve efficiency then a $3\times3$ convolution layer follows to extract new features. Next, the second and the third Downsampler Blocks are replaced by the transition layers proposed in DenseNet, which are made up of a $1\times1$ convolution layer followed by a $2\times2$ average pooling layer. Since extracting depth features cannot benefit by simply using a deeper network, we only place 3 and 4 dense modules in the second and the third dense block, respectively to save computational cost. Instead, we employ a larger growth rate for each dense module to make D&Y Encoder wider. This shallow but wide design is able to improve efficiency with little performance degradation in our case.

On top of that, to fully make use of the depth information, we add a dense block in a shallow position called Shallow

Block right after the first Downsampler Block to extract more boundary information for efficaciously addressing the object localization issue in semantic segmentation. The benefits of Shallow Block will be shown in Section 3.3.

In order to reduce the defects of the captured depth maps used by D&Y Encoder, we stack the luminance images with the depth maps as two-channel inputs. The luminance information can guide D&Y Encoder to suppress the noise effects contained in the depth maps and extract valid information for segmentation.

### 2.3. Fusion Mechanism

We take the essence of the fusion idea introduced in FuseNet [6] and further develop a more effective approach in our fusion-based LDFNet. According to FuseNet and our experimental results, a simple four channels stack cannot effectively extract information from the depth map. Hence, instead of simply appending the depth channel to the RGB channels, we adopt the design of two parallel sub-networks. However, different from FuseNet that simply uses an identical structure for both the main RGB sub-network and the D&Y sub-network, our network adopts different architectures for them. The output features of each dense block in D&Y Encoder is fused to RGB Encoder at the same resolution by the element-wise summation (see Figure 2). We also fuse the features after each transition layer, so there are five fusion operations in total. To allow this fusion process, the difference in the numbers of channels in the two encoders is eliminated by using properly the 1×1 convolution layers. Our fusion mechanism enables our network to integrate the multi-modal information in an efficient manner.

## 3. EXPERIEMENTS

In this section, we conduct a series of experiments to evaluate the effectiveness of our network design choices and compare its performance with other schemes.

### 3.1. Implementation Details

Our networks are trained by using Adam optimization [9]. The L2 weight decay of the optimizer is set to 0.0001, and the batch size is set to 4. Also, due to the imbalance of pixels of each class presented in the dataset, a classical class weighting scheme defined in [13] is employed: $\omega_{class} = 1/\ln(c + p_{class})$, where we set $c$ to 1.1 in our case. The initial learning rate is 0.0005, and the poly learning rate policy [2] is used. We also include the dropout layers [15] at the end of each Non-Bottleneck and dense module in training with a rate of 0.05 as regularization. For our dense blocks, we set the growth rate to 42. Every convolutional layers are followed by a batch normalization layer [8] and a ReLU. We also adopt data augmentation in training by using random horizontal flip and a translation of 0~2 pixels on both axes. The mean of intersection-over-union (mIoU) is the evaluation metric.

### 3.2. Dataset

We use Cityscapes dataset [3], which consists of 5,000 pixel-level finely annotated street scene images. The overall dataset is divided into three subsets: training, validation, and testing with 2,975, 500 and 1525 images, respectively. Totally, 19 classes such as building, road, and pedestrian are defined in the Cityscapes dataset. The testing data labels are unavailable, but we can evaluate our network on the online test server. The original dataset resolution is 1024×2048 and they are resized to 512×1024 for our training process.

### 3.3. Ablation Study

We vary the network structure to see the performance of different network design choices. The experimental results are summarized in Table 1.

First, ERFNet-Depth uses only the depth maps for prediction. The result indicates that the depth maps can provide a certain amount of information for this purpose, but its accuracy is low, compared to the RGB images. Then, we try two structures to process the depth information: 1) stacking the depth maps as the 4$^{th}$ input channel, and 2) using a two-branch architecture. ERFNet-RGB uses the RGB input images only. ERFNet-Stack that simply stacks RGB and D channels produces similar results as ERFNet-RGB. In other words, the stack method cannot benefit from the additional depth information. By contrast, the proposed LDFNet achieves a significant improvement, a mIoU of 68.33%. The difference of the mIoU scores between our method and ERFNet-Stack shows that our fusion mechanism is a more effective design for depth information extraction. Proper use of the depth map can boost accuracy.

Next, we examine the capability of different structures in constructing D&Y Encoder. Compared to LDFNet, LDF-non-Dense uses the ERFNet-based [14] structure; that is, its D&Y Encoder is identical to RGB Encoder. The results show that LDFNet can obtain a higher mIoU score with fewer parameters. Therefore, adopting the dense connectivity [4] is a preferred solution.

We next confirm the advantages of using Shallow Block, which is located after the first Downsampler Block in the D&Y Encoder. Both LDF-w/o-Shallow and LDF-58-w/o-Shallow discard Shallow Block, but LDF-58-w/o-Shallow increases the numbers of dense modules in its second and third dense blocks to 5 and 8 respectively. Compared to these two, LDFNet can achieve higher accuracy, even though LDF-58-w/o-Shallow has more modules in its deeper layers. Our reasoning is that the depth information has a strong correlation to the object edge, contour, and boundary information, so placing Shallow Block at the early stage is beneficial to extract these desired low-level features.

Furthermore, we would like to show the usefulness of using the luminance information in D&Y Encoder. Because the depth maps produced by depth sensors like Kinect contain defects and the resolution of depth sensors is relatively small compared to its RGB counterpart, these sensor errors would lead to incorrect information be fused into RGB Encoder. After inserting the luminance information into the depth processing branch, the noise effects could be suppressed. The experimental results verify this conjecture; that is, comparing LDF-w/o-Y that does not use luminance information to LDFNet, there is a great improvement in the mIoU score.

Table 1: Evaluation results on the Cityscapes validation set, comparing the proposed LDFNet with different design choices.

| Method | RGB Inputs | Depth Maps | Y Info. | Shallow Block | Dense Connects | mIoU (%) | Parameters |
|---|---|---|---|---|---|---|---|
| ERFNet-Depth |  | ● |  |  |  | 47.48 | 1.97M |
| ERFNet-RGB | ● |  |  |  |  | 65.59 | 1.97M |
| ERFNet-Stack | ● | ● |  |  |  | 65.06 | 1.97M |
| LDF-non-Dense | ● | ● | ● |  |  | 66.53 | 2.95 M |
| LDF-w/o-Shallow | ● | ● | ● |  | ● | 66.54 | 2.20 M |
| LDF-58-w/o-Shallow | ● | ● | ● |  | ● | 65.93 | 2.42 M |
| LDF-w/o-Y | ● | ● |  | ● | ● | 65.72 | 2.31M |
| LDF-RGB-RGB | ● |  |  | ● | ● | 67.79 | 2.31M |
| LDFNet | ● | ● | ● | ● | ● | 68.48 | 2.31M |

Table 2: Evaluation results on the Cityscapes test set, comparing LDFNet with the other RGB-D methods.

| Method | mIoU (%) | Speed (fps) |
|---|---|---|
| MultiBoost | 59.3 | 4.0 |
| Pixel-level Encoding [16] | 64.3 | n/a |
| Scale invariant CNN+CRF [10] | 66.3 | n/a |
| RGB-D FCN | 67.4 | n/a |
| LDFNet (ours) | 71.3 | 18.4 |

Table 3: Comparison of model efficiency with RGB methods. Sub: the amount of subsampling used by the method at test time.

| Method | Parameters | Sub | Speed (fps) |
|---|---|---|---|
| DeepLabv2 [2] | 44.0M | no | n/a |
| PSPNet [20] | 65.7M | no | n/a |
| Dilation10 [19] | 140.8M | no | 0.25 |
| FCN-8s [12] | 134.5M | no | 2.0 |
| SegNet [1] | 29.5M | 4 | 16.7 |
| LDFNet (ours) | 2.31M | 2 | 18.4 |

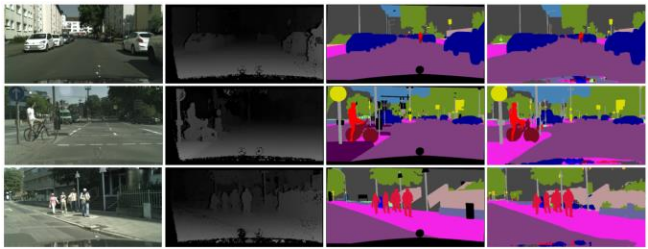

Figure 3: Sample results of LDFNet on the Cityscapes validation set. From left to right: (a) RGB image, (b) depth map, (c) Ground truth, (d) LDFNet.

Finally, although LDFNet has only slightly more parameters than its backbone model, ERFNet, we would like to testify whether the improvement is coming from the proposed fusion mechanism or simply due to the increased parameters. Thus, we build LDF-RGB-RGB, which is identical to the LDFNet structure except that its inputs are two duplicate RGB images fed into the two branches respectively. Its accuracy is between ERF-RGB and LDFNet, demonstrating the increased parameters indeed provide some improvements, but our fusion mechanism of incorporation multi-modal information contributes significantly more.

### 3.4. Evaluation Results

We train LDFNet in two stages (both the training and validation data are included in training) for the final evaluation. First, we train only the two encoders by downsized labels. Second, we add the decoder together with the encoders in training. We do not use any testing tricks such as multi-crop and multi-scale testing in evaluations. In Table 2, we report the results evaluated on the Cityscapes test set and the comparisons with the other state-of-art systems. LDFNet achieves a 71.3% mIoU score without any pretrained model and surpasses all the other methods [10,16] designed for RGB-D semantic segmentation on this benchmark.

Moreover, in Table 3, LDFNet outperforms several state-of-art networks for the RGB semantic segmentation task in terms of efficiency, such as DeepLab [2] and PSPNet [20]. Even though LDFNet processes the extra depth information, the entire network has fewer parameters and maintains a faster inference speed. LDFNet can run on the resolution 512×1024 inputs at the speed of 18.4 and 27.7 frames per second (fps) on a single Titan X Maxwell and GTX 1080Ti respectively. Some visual results are shown in Figure 3.

### 4. CONCLUSION

In this study, we propose a novel information-fused network, LDFNet, to incorporate luminance, depth, and color information for RGB-D semantic segmentation. LDFNet is able to effectively extract the features from both the RGB images and the depth maps to achieve a higher segmentation performance, while it maintains a rather low computational complexity. After conducting a series of experiments, we demonstrate the effectiveness of our design choices. LDFNet successfully outperforms the other state-of-the-art systems on an influential benchmark.